\documentclass{article}
\usepackage{spconf,amsmath,graphicx}
\usepackage{amsfonts}
\usepackage{amsmath}
\usepackage{adjustbox}
\usepackage{algorithmicx}
\usepackage{algpseudocode}
\usepackage[ruled]{algorithm}
\usepackage{graphicx}
\usepackage{caption}
\usepackage{graphicx,times,amsmath} % Add all your packages here
\usepackage{caption}
\usepackage{subcaption}
\usepackage[rightcaption]{sidecap}
\usepackage{caption}
\usepackage{multirow}
\usepackage{hhline}
\usepackage{amsmath}
\usepackage{epstopdf}
\usepackage{tabularx}

\usepackage{amsfonts}
\usepackage{amsmath}
\usepackage{adjustbox}
\usepackage{algorithmicx}
\usepackage{algpseudocode}
\usepackage[ruled]{algorithm}
\usepackage{graphicx}
\usepackage{caption}
\usepackage{graphicx,times,amsmath} % Add all your packages here
\usepackage{makecell}

\usepackage{caption}
\usepackage{subcaption}
\usepackage[rightcaption]{sidecap}
\usepackage{caption}
\usepackage{multirow}
\usepackage{hhline}
\usepackage{amsmath}
\usepackage{epstopdf}
\usepackage{tabularx}
\usepackage{algorithmicx}
\usepackage{algpseudocode}
\usepackage[ruled]{algorithm}
\usepackage[labelformat=simple]{subcaption}

\usepackage{lipsum}
\usepackage{fancyhdr}

\makeatletter
\def\old@comma{,}
\catcode`\,=13
\def,{%
  \ifmmode%
    \old@comma\discretionary{}{}{}%
  \else%
    \old@comma%
  \fi%
}
\makeatother

\graphicspath{ {images/} }

\usepackage{alphalph}
\title{Cylindrical Transform: 3D Semantic \\Segmentation of Kidneys With Limited Annotated Images}

\name{Hojjat~Salehinejad$^{\star \dagger}$, Sumeya Naqvi$^{\dagger}$, Errol Colak$^{\dagger}$,~Joseph~Barfett$^{\dagger}$, and Shahrokh~Valaee$^{\star}$\thanks{The authors thank the support of NVIDIA Corporation.}}
\address{$^{\star}$Department of Electrical \& Computer Engineering, University of Toronto, Toronto, Canada \\
$^{\dagger}$Department of Medical Imaging, St. Michael's Hospital, University of Toronto, Toronto, Canada \\
\textit{hojjat.salehinejad@mail.utoronto.ca, \{naqvis,colake,barfettj\}@smh.ca, valaee@ece.utoronto.ca}}

\begin{document}

\newcommand*{\img}{%
  \includegraphics[
    width=\linewidth,
    height=20pt,
    keepaspectratio=false,
  ]{example-image-a}%
}

\maketitle
\thispagestyle{fancy}

\begin{abstract}
In this paper, we propose a novel technique for sampling sequential images using a cylindrical transform in a cylindrical coordinate system for kidney semantic segmentation in abdominal computed tomography (CT). The images generated from a cylindrical transform augment a limited annotated set of images in three dimensions. This approach enables us to train contemporary classification deep convolutional neural networks (DCNNs) instead of fully convolutional networks (FCNs) for semantic segmentation. Typical semantic segmentation models segment a sequential set of images (e.g. CT or video) by segmenting each image independently. However, the proposed method not only considers the spatial dependency in the x-y plane, but also the spatial sequential dependency along the z-axis. The results show that classification DCNNs, trained on cylindrical transformed images, can achieve a higher segmentation performance value than FCNs using a limited number of annotated images.

\end{abstract}



\section{Introduction}
Object recognition refers to the task of detecting and labeling all objects in a given image. A bounding box is usually used in this approach to localize the object(s). 
In object detection, bounding boxes are used to localize a specific object in the image and the rest of the image is assigned to the non-object class. Semantic segmentation refers to the classification of each pixel in an image to generate an image mask consisting of a number of labeled regions. Object recognition approaches are generally easier to implement and computationally less expensive than semantic segmentation methods. However, accuracy and pixel-depth segmentation can be more important than computational complexity in certain applications such as  medical image processing.

Deep fully convolutional networks (FCNs)~\cite{long2015fully} are popular models for semantic segmentation~\cite{sharma2017automatic} that use a convolutional decoder with a large annotated training dataset. Since there are limited numbers of annotated images and data samples available for every possible class in real-world problems~\cite{pouladi2015recurrent}, augmentation methods such as image rotation and synthesis~\cite{salehinejad2018generalization} can help increase the diversity of training datasets, and therefore prevent the models from overfitting~\cite{sharma2017automatic}, \cite{salehinejad2018image}, \cite{salehinejad2017recent}. In \cite{salehinejad2018image}, we have proposed a radial transform method in the polar coordinate system as a novel augmentation method for classification problems. This technique is well suited for highly imbalanced datasets, or datasets with a limited number of labeled images. 

In this paper, we propose a cylindrical transform in the cylindrical coordinate system as a technique to generate representations from limited annotated sequential images. The cylindrical transform method enables us to train contemporary classification deep convolutional neural networks (DCNNs) instead of FCNs for semantic segmentation. We applied the proposed method for registration-free segmentation of left kidney, right kidney, and non-kidney data classes in abdominal computed tomography (CT) images by training AlexNet~\cite{krizhevsky2012imagenet} and  GoogLeNet~\cite{szegedy2015going} DCNNs. We have selected these architectures due to their simplicity of training and relatively high classification performance~\cite{canziani2016analysis}.  

%We achieved an overall mean Dice similarity coefficient (DSC)~\cite{sharma2017automatic} of $91.91\%$ by using 378,000 cylindrical transform generated images with a sampling rate of 1,000 per class per image from only 7 abdominal CTs, for a total of 126 original annotated axial images (18 images per CT). 

\section{Proposed Method}
In this section, we discuss the proposed cylindrical transform
sampling method and the training and inference
procedures of a DCNN using cylindrical transform generated images for
semantic segmentation.

\subsection{Sampling Using Cylindrical Transform in 3D Space}
\begin{figure}[!tp]
\centering
\captionsetup{font=small}
                \includegraphics[width=0.3\textwidth]{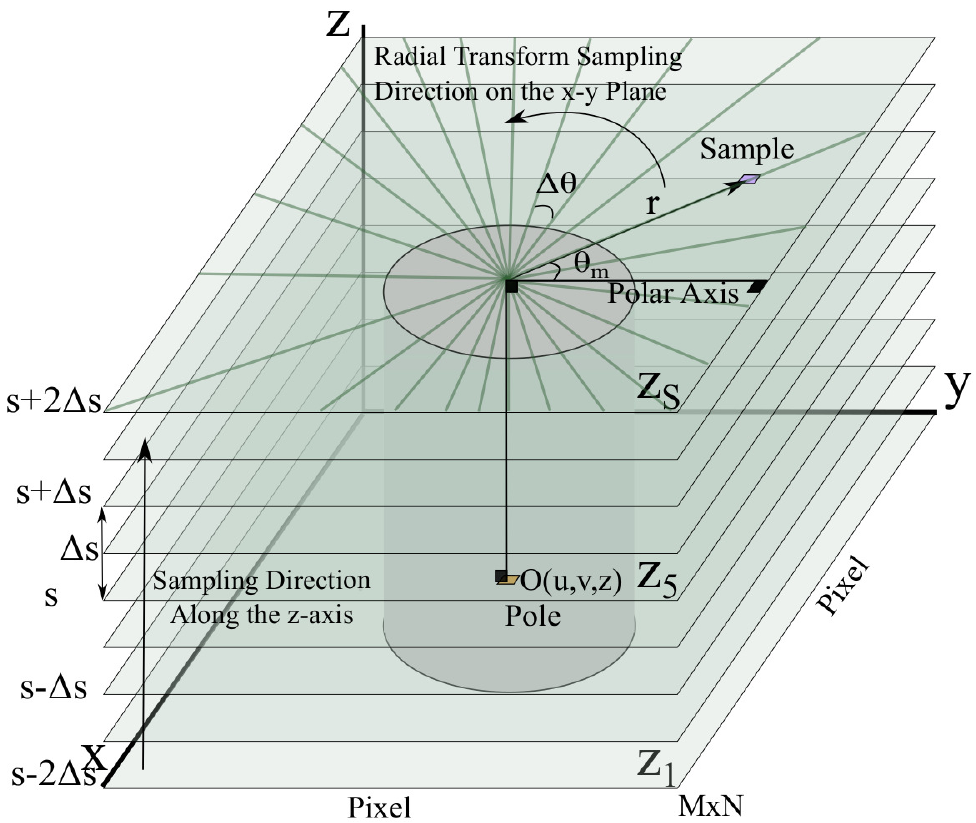}  
        \caption{Sampling a sequence of images using cylindrical transform.}
        \label{fig:cyledrical_sampling_model}
        \vspace{-1mm}
\end{figure}

A cylindrical coordinate system is a generalization of a polar coordinate system to  3D space and is created by superposing a height along the z-axis. The objects in a volume of images not only have spatial dependency on the x-y plane, but also along the z-axis. We define a volume $\mathbf{X}$ as a sequence of $S$ images $\mathbf{X}_{s}\: \forall s\in\{1,...,S\}$ along the z-axis, where $\mathbf{X}_{s}$ is of size $M\times N$, as presented in  Figure~\ref{fig:cyledrical_sampling_model}. We can randomly select a pixel from $\mathbf{X}$ in the Cartesian coordinate system, such as ${(m,n,s)\in\mathcal{C}^{3}}$. This pixel can be mapped onto the cylindrical coordinate system as a pole with coordinates $O(u,v,z)\in\mathcal{P}^{3}$. Cylindrical transform represents each pixel in the
volume $\mathbf{X}$ as a new image of the size $S^{'}\times N$, where $S^{'}=S\times M$, by up-sampling the
pole and representing the spatial information
between the pole and other pixels in the volume.

\alglanguage{pseudocode}
\begin{algorithm}[!t]
%\textbf{Algorithm 2: Sampling Using Spherical Transform in 3D Space}
\caption{Sampling via Cylindrical Transform}
\scriptsize

\begin{algorithmic}%[1]
\State Read $O(u,v,z)$ // Selected pole in the image volume
\State Read $\mathbf{X}$ // Original image volume
\State Read $\Delta s$ // Sampling step along slices
\State Initialize $\mathbf{\hat{X}}$ to zero // Cylindrical transform image of size $(S^{'}\times M)\times N$ 

\For {$j=0 \to (S^{'}-1)/2$}
\If {$0\leq z\pm j\cdot \Delta s < S$}
\State $\Xi(j)= z\pm j\cdot \Delta s$ // Set of slices to sample
\EndIf
\EndFor

\For {$s=0 \to S^{'}-1$}
\For {$m=0 \to M-1$}
\State $\theta_{m}=2\pi m/M$
\For {$r=0 \to N-1$}
\State $\hat{x}=round(r\cdot cos(\theta_{m}))$
\State $\hat{y}=round(r\cdot sin(-\theta_{m}))$
\State $\hat{z}=\Xi(s)$
\If {$0\leq u+\hat{x} <M \:\&\: 0\leq v+\hat{y} <N$} 
\State $n=r$
\State $\mathbf{\hat{X}}(s\cdot M+m,n)=\mathbf{X}(u+\hat{x},v+\hat{y},\hat{z})$
\EndIf
\EndFor
\EndFor
\EndFor

\end{algorithmic}
\label{alg:cylindrical}
\end{algorithm}
\alglanguage{pseudocode}

In the cylindrical coordinate system, a pixel on the plane $z$ can be represented as $(r,\theta,z)\in \mathcal{P}^{3}$, where $r\in \mathbb{Z}^{+}$ is the radial coordinate from the pole along the z-axis and $\theta\in \mathbb{R}^{+}$ is the counter-clockwise angular coordinate. It is considered with respect to an axis drawn horizontally from the pole to the right, as illustrated in Figure~\ref{fig:cyledrical_sampling_model}. For a given volume of images, we can select $(S^{'}-1)/2$ slices, with a given distance of $\Delta s$ slices, above and under the slice $\mathbf{X}_{z}$ with respect to the pole $O$ such as ${\Xi=\cup_{j=0}^{(S^{'}-1)/2}\{z\pm j\cdot \Delta s | 0\leq z\pm j\cdot \Delta s < S\}}$. In the cylindrical coordinate system $\mathcal{P}^{3}$, we can generate ${K=\cup_{k=0}^{(S^{'}\times M)\times N-1}\{(r,\theta_{m},z)_{k}\}}$ sampling points with respect to a pole $O(u,v,z)$ such that
\begin{equation}
\theta_{m} = 2\:\pi \cdot m/M,
\end{equation}
for $m\in\{0,...,M-1\}$. By $\textbf{X} \xrightarrow{\phi (r_{n},\theta_{m},z)_{k}} \hat{\textbf{X}}$, we project the pixels at Cartesian coordinates $(m,n,s)_{k}\in \mathbb{Z}^{+}$ from the original image $\textbf{X}\in\mathcal{C}^{3}$ to generate an image $\hat{\textbf{X}}\in \mathcal{C}^{3}$ with respect to the pole using cylindrical transform $\phi(\cdot)$ as
\begin{equation}
\hat{x} = round(r_{n} \cdot cos(\theta_{m})) \:\:  \& \:\:  \hat{y} = round(r_{n} \cdot sin(-\theta_{m}))
\label{eq2}
\end{equation}
for $r_{n}\in \{0,...,N-1\}$ and $m\in \{0,...,M-1\}$ such that~${0\leq u+\hat{x} < M}$, ${0 \leq  v+\hat{y} < N}$, and $round(\cdot)$ is the rounding function to the nearest integer. These conditions guarantee that the pair $(\hat{x},\hat{y},\hat{z})$ stays spatially within $\hat{\textbf{X}}$. 
A pixel $(\hat{m},\hat{n})$ in the constructed image is then defined as ${\hat{x}_{\hat{m},\hat{n}}=x_{u+\hat{x}, v+\hat{y},z}}.$
The image $\hat{\textbf{X}}$ is the cylindrical transform image of $\textbf{X}$ with respect to the pole $O(u,v,z)\in \textbf{X}$ with sampling step of $\Delta s$ along the z-axis. Algorithm~\ref{alg:cylindrical} shows the pseudocode of the cylindrical transform sampling.

\begin{figure}[!tp]
\centering
\captionsetup{font=small}     
          \begin{subfigure}[t]{0.17\textwidth}
        \centering
                \includegraphics[width=1\textwidth]{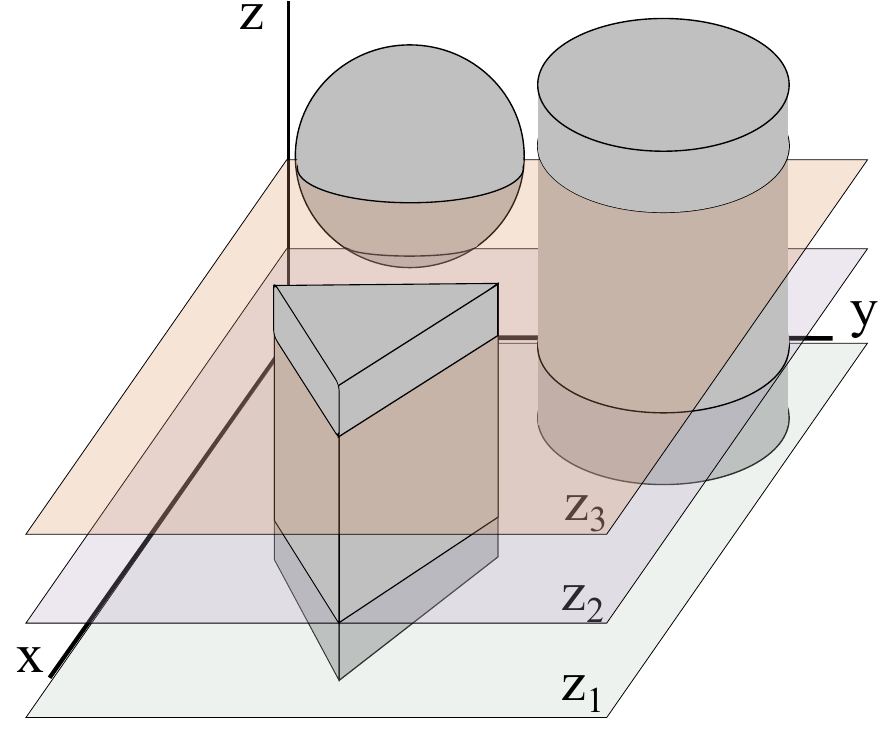}
                \caption{Geometric shapes. }
                \label{fig:objects_raw}
        \end{subfigure}%          
                      \begin{subfigure}[t]{0.32\textwidth}
        \centering
                \includegraphics[width=1\textwidth]{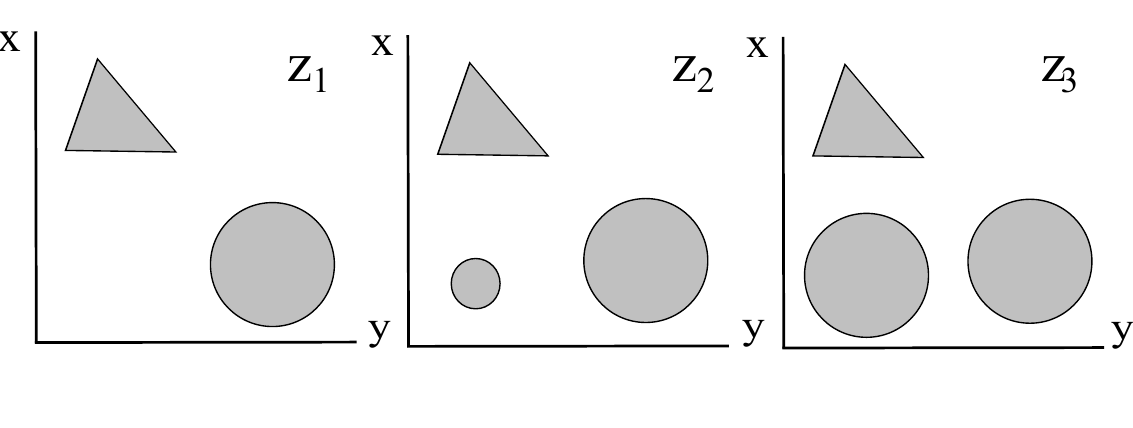}
                \caption{Slice view of the geometric shapes. }
        \end{subfigure}%      
        \vspace{2mm}
                             \begin{subfigure}[t]{0.49\textwidth}
        \centering
                \includegraphics[width=0.7\textwidth]{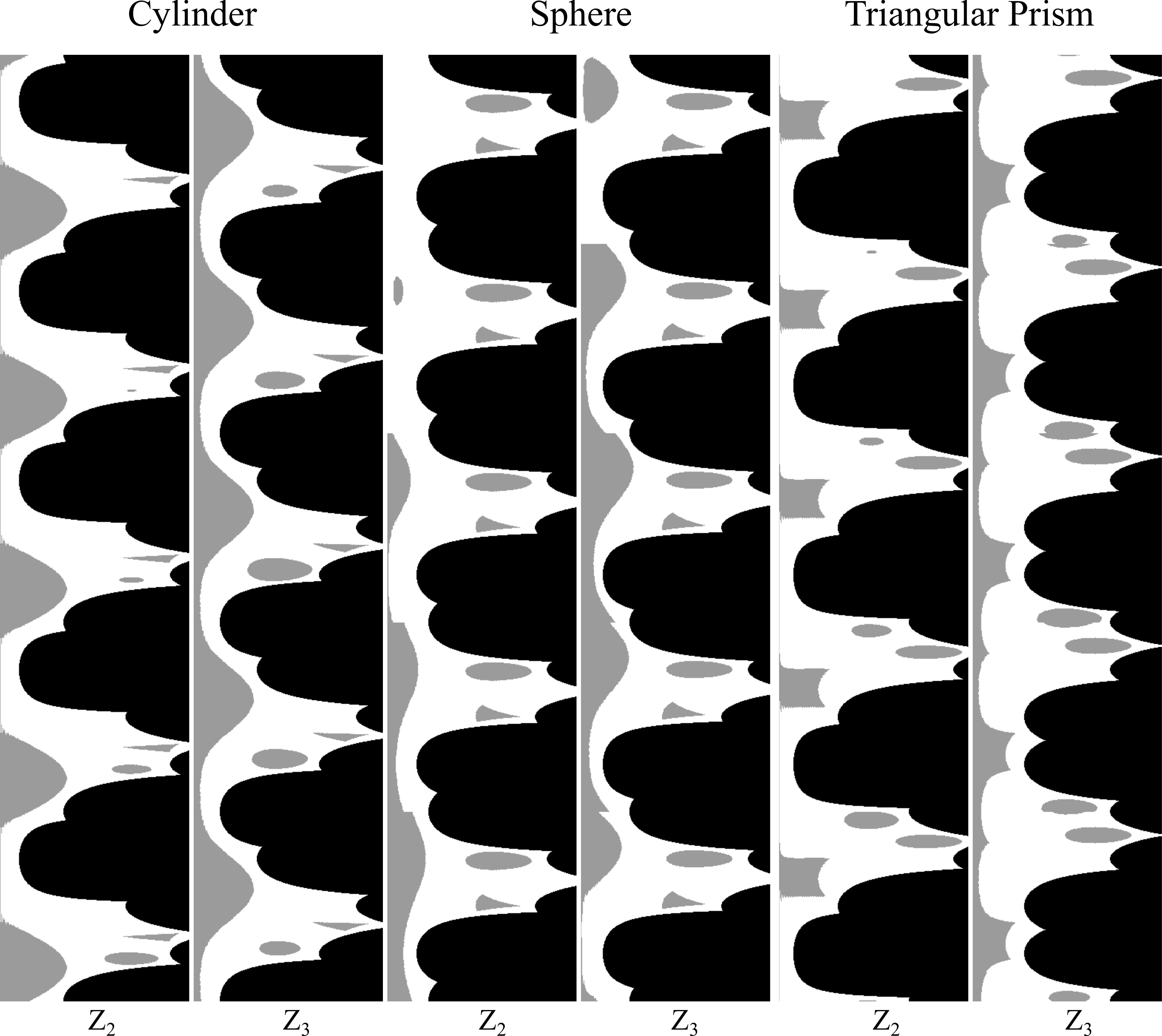}
 \caption{Cylindrical transform of randomly selected poles from (a). The slice number along the z-axis are ${\mathrm{z}_{1}=0; \mathrm{z}_{2}=63; \mathrm{z}_{3}=90}$. }
        \end{subfigure}%       
        
        \caption{A volume of geometric shapes in a 3D Cartesian space. The pole is selected randomly inside each geometric shape. In $z_{3}$, the sphere and cylinder are presented as two identical circles, while they belong to two different objects in the volume. However, $z_{2}$ shows the difference of these objects at a different spatial location along the z-axis. Cylindrical transform provides a representation of the objects by considering the spatial dependency on the x-y plane as well as the z-axis. This observation is not detectable by only considering a single slice along the x-y plane.}
\label{fig:objects}
\end{figure}

Figure~\ref{fig:objects} shows the advantage of using cylindrical transformed images over independent slices of a volume. An sphere and cylinder look different in a 3D space. However, these objects may look similar depending on the object location along the z-axis on the x-y plane. This is while the cylindrical transformed images capture the spatial difference along the z-axis and by combining that with spatial information on the x-y plane, represent a volume along an arbitrary pixel as an image, feasible for machine learning. This image contains information about spatial dependency on the x-y plane as well as the z-axis.

\subsection{Cylindrical Transform for Semantic Segmentation}
\begin{figure}[!tp]
\centering
\captionsetup{font=small}     
          \begin{subfigure}[t]{0.49\textwidth}
        \centering
                \includegraphics[width=1\textwidth]{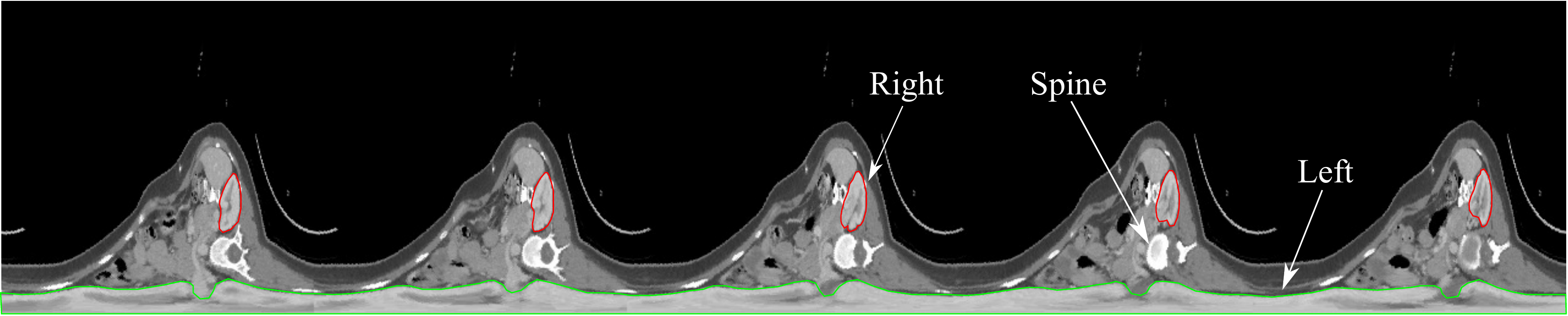}
                \caption{Left kidney. }
                \label{fig:objects_raw}
        \end{subfigure}%          
        
                      \begin{subfigure}[t]{0.49\textwidth}
        \centering
                \includegraphics[width=1\textwidth]{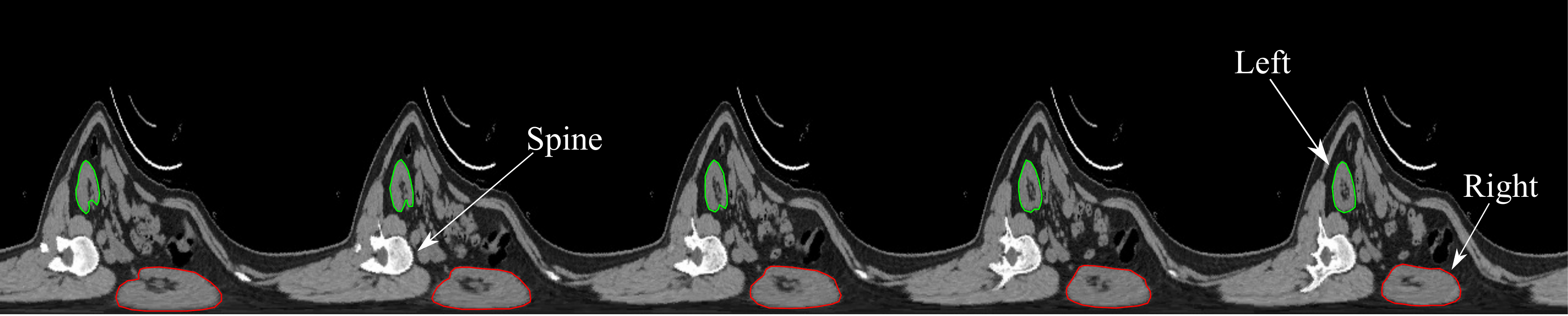}
                \caption{Non-kidney.}
                \label{fig:slices}
        \end{subfigure}%    
             
                             \begin{subfigure}[t]{0.49\textwidth}
        \centering
                \includegraphics[width=1\textwidth]{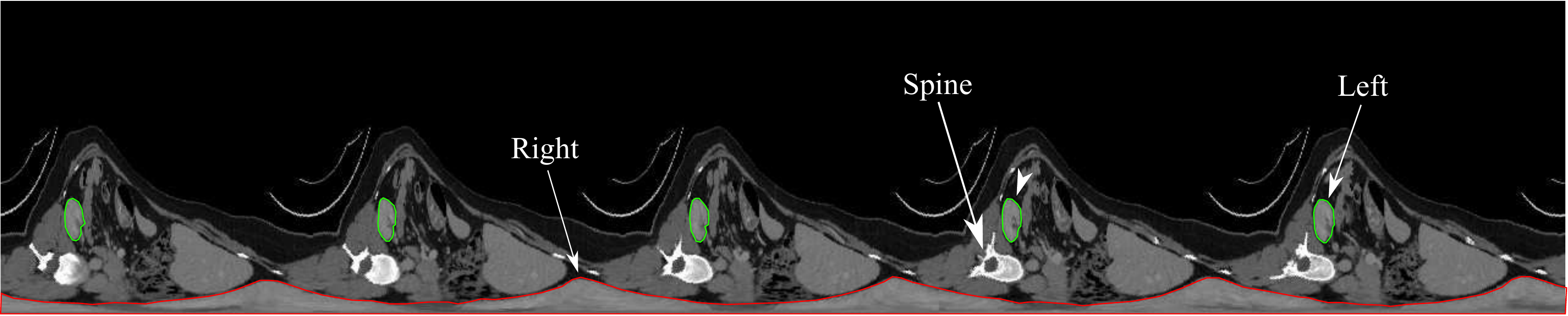}
 \caption{Right kidney}                \label{fig:slices}
        \end{subfigure}%       
     \vspace{-3mm}   
        \caption{Samples of cylindrical transform generated images for left kidney, non-kidney, and right kidney. The images are rotated $90^{o}$ counter-clockwise for the sake of presentation.}
\label{fig:kidney_cylendrical_samples}
\end{figure}

\begin{figure}[!tp]
\centering
\captionsetup{font=small}
                \includegraphics[width=0.49\textwidth]{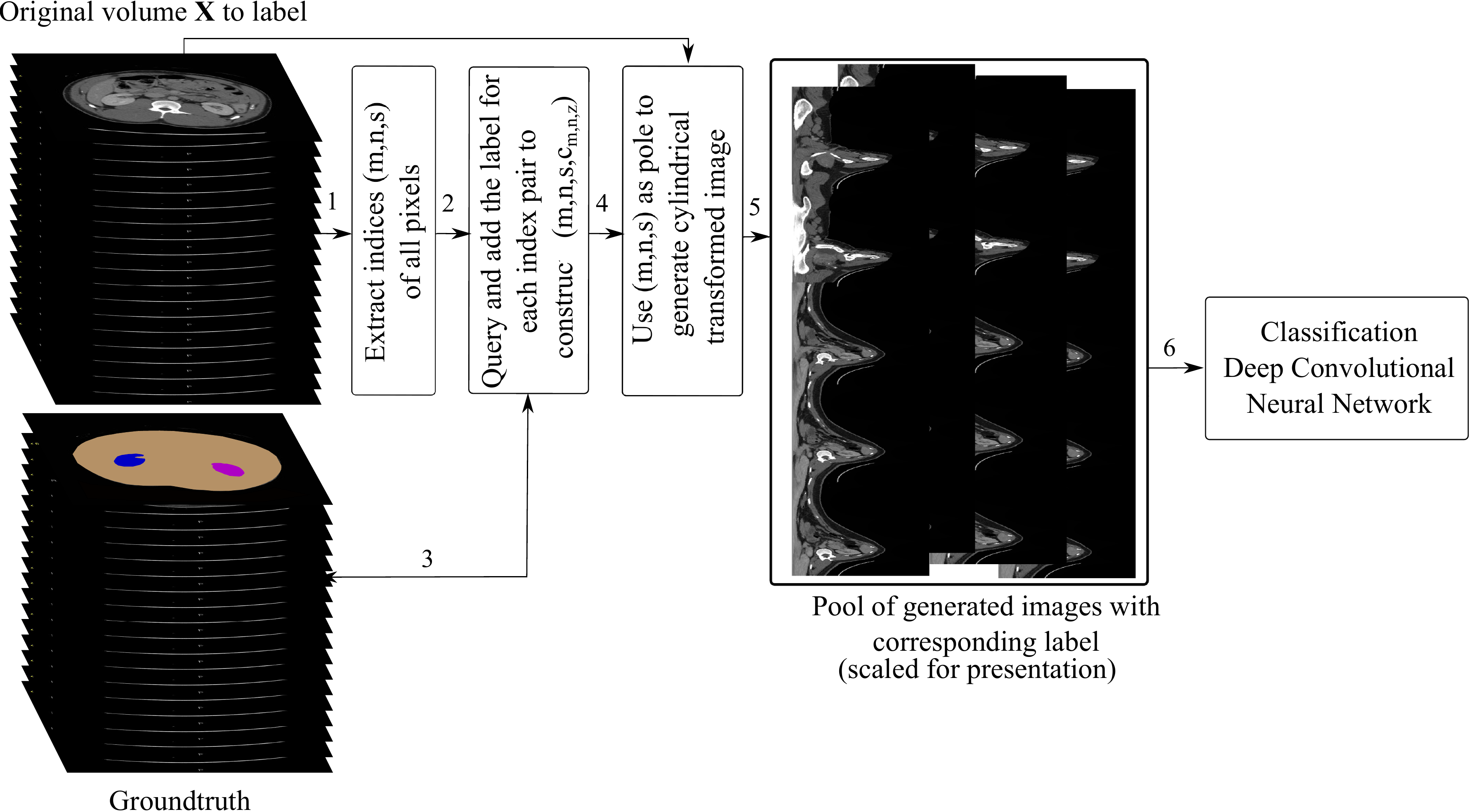}  
        \caption{Training semantic segmentation using classification DCNNs with cylindrical transformed images. The steps are labeled.}
   \vspace{-4mm}     
\label{fig:cylendrical_training_model}
\end{figure}

Figure~\ref{fig:kidney_cylendrical_samples} shows samples of cylindrical transform generated images from contrast-enhanced abdominal CT. Figure~\ref{fig:cylendrical_training_model} shows the
procedure for training a DCNN with cylindrical transformed images. 
By considering a sequence of images $\mathbf{X}$ as the input, the cylindrical
transform generates images for a number of
randomly selected poles in $\mathbf{X}$ and stores them with their
corresponding labels in a pool of images to train a DCNN. The trained model can later be used
for inference, where the cylindrical transform
considers every pixel in the original image $\mathbf{X}$ as the pole
and generates its corresponding cylindrical transform image. The
generated images are then passed to the trained DCNN for
classification and labeling of a mask template, which represents
the predicted data class for each pixel in $\mathbf{X}$.

\section{Experiments}
%In this section, we discuss the results of our experiments for 
%semantic segmentation of left kidney, right kidney, and non-kidney pixels in contrast-enhanced abdominal CTs. 

\subsection{Data}
%from our medical institution's picture archiving and communication system (PACS) 
With the approval of the research ethics board, 20 contrast-enhanced normal abdominal CT acquisitions from an equal number of male and female subjects between 25 to 50 years of age were collected~\cite{salehinejad2017interpretation}. Each acquisition had on average 18 axial slices containing kidneys.
The left and the right kidneys were outlined manually by trained personnel and stored as $256\times256$ images. The boundary delineation was performed using a standard protocol for all kidneys. To avoid inter-rater variability in the dataset, quality of segmentation was assured by two board certified radiologists.
The sampling step along the z-axis is $\Delta s=3$ and the size of a cylindrical transform generated image is $(S^{'}\times M)\times N=(5\times 256)\times 256$. 

\subsection{Technical Details of Training}
The FCN models were trained on $7\times 18=126$ original images with a setup as outlined in~\cite{sharma2017automatic}. For experiments with cylindrical transformed images, 7 acquisitions each containing on average 18 axial slices (totalling $7\times 18=126$) were used for training with 1,000 randomly selected poles per label class per slice to generate cylindrical transformed images. For all experiments, three acquisitions were used for validation ($3\times18=54$ axial slices), and 10 acquisitions ($10\times18=180$ axial slices) were used for test. The number of training iterations was set to 120. An Adam~\cite{kingma2014adam} optimizer with a sigmoid decay adaptive learning rate (LR) and momentum term of 0.9 was used. The activation function before the max-pooling layer was a ReLU~\cite{nair2010rectified}. The $L_{2}$ regularization was set to $1\times 10^{-4}$ and early-stopping (storing network parameters and stopping at maximum validation performance in a window of 5 iterations) was applied.
The training datasets were shuffled in each training epoch. The performance results were collected after 10-fold cross-validation.

\subsection{Semantic Segmentation of Kidneys in Contrast-Enhanced Abdominal Computed Tomography}
%For a semantic segmentation task, it is necessary to apply a label to every pixel of an image. Therefore, radial transform needs to be applied to each pixel of a give image for classification.
Using the definition of true positive ($T_{P}$), false positive ($F_{P}$), and false negative ($F_{N}$), the precision ${P=T_{p}/(T_{p}+F_{p})}$ and recall ${R=T_{p}/(T_{p}+F_{n})}$ measure the success of prediction in classification tasks~\cite{davis2006relationship}. The Dice similarity coefficient (DSC)~\cite{sharma2017automatic} is a well-known measure for the accuracy of segmentation methods~\cite{sharma2017automatic}. By considering a volume as a set of pixels, for a segmented sequence of images $\tilde{\mathbf{X}}$ and its corresponding ground-truth $\bar{\mathbf{X}}$, the DSC is expressed as
${DSC(\tilde{\mathbf{X}},\bar{\mathbf{X}})=2|\tilde{\mathbf{X}}\cap \bar{\mathbf{X}}|/(|\tilde{\mathbf{X}}|+|\bar{\mathbf{X}}|)}$, where $|\cdot|$ is the cardinality of the set. Since we apply the transform to each pixel of the volume $\mathbf{X}$, the DSC segmentation accuracy can be interpreted as the top-1 classification accuracy~\cite{lapin2016loss}.

\begin{table}[]
\centering
\captionsetup{font=footnotesize}
\caption{DSC value of classification DCNNs trained on cylindrical transform generated images and FCNs for semantic segmentation of kidneys in contrast-enhanced abdominal CTs. CLT: cylindrical transform; P: pre-trained on ImageNet; LR: adaptive learning rate basis; MB: size of mini-batch. Top DSC value is in boldface.}
\begin{adjustbox}{width=0.49\textwidth}

\begin{tabular}{|c|c|c|c|c|c|c|}
\hline
\multirow{5}{*}{Model} &\multirow{5}{*}{\rotatebox[origin=c]{90}{LR} } &\multirow{5}{*}{\rotatebox[origin=c]{90}{MB}}& \multicolumn{3}{c|}{{\begin{tabular}[c]{@{}c@{}}DSC per class\end{tabular}} } & \multirow{5}{*}{\makecell{DSC}} \\ \cline{4-6}
  &  &&\rotatebox[origin=c]{90}{Left Kidney}  & \rotatebox[origin=c]{90}{Non-Kidney} & \rotatebox[origin=c]{90}{Right Kidney}    &  \\ \hline \hline
 
%FCN-AlexNet 	&	 36.72\% 	&	 35.53\% 	&	 37.66\% 	&	 36.63\%  \\ \cline{1-1} \cline{2-5}

FCN-GoogLeNet 	& $10e^{-5}$&64&	 42.17\% 	&	 43.73\% 	&	 40.36\% 	&	 42.08\%  \\ \hline
FCN-VGG-19	&	$10e^{-5}$& 64&	34.52\%	&	 35.93\% 	&	 36.26\% 	&	 35.57\% \\ \hline
FCN-VGG-19P	&	$10e^{-5}$& 64&	54.26\%	&	 58.93\% 	&	 56.27\% 	&	 56.48\% \\ \hline \hline
%FCN-AlexNet (augmented)	&	55.32\%	&	 60.20\% 	&	 57.57\% 	&	 57.69\% \\ \cline{1-1} \cline{2-7}
FCN-GoogLeNet (augmented)	& $10e^{-5}$&64&	60.62\%	&	 65.82\% 	&	 62.69\% 	&	 63.04\% \\ \hline
FCN-VGG-19 (augmented)	&	$10e^{-5}$& 64&	62.74\%	&	65.20\% 	&	 63.36\% 	&	 63.76\% \\ \hline
FCN-VGG-19P (augmented)	&	$10e^{-5}$& 64&	66.83\%	&	69.26\% 	&	 67.77\% 	&	 67.95\% \\ \hline\hline

CLT-AlexNet 	&	$10e^{-5}$& 4& 93.68\% 	&	92.55 \% 	&	94.42 \% 	&	 93.52\%    \\ \cline{1-1} \cline{2-7}
CLT-GoogLeNet 	&$10e^{-5}$	&4 &$\mathbf{98.00\%}$ 	&	 $\mathbf{97.80\%}$ 	&	 $\mathbf{99.47\%}$ 	&	 $\mathbf{98.40\%}$  \\ \cline{1-1} \cline{2-7}
%RT-VGG-19 	&$10e^{-5}$	&64 & \% 	&	  \% 	&	  \% 	&	  \%  \\ \cline{1-1} \cline{2-7}
%RT-VGG-19P 	&	$10e^{-5}$& 64&  \% 	&	  \% 	&	 \% 	&	 \%    \\ \cline{1-1} \cline{2-7}
\end{tabular}
\end{adjustbox}
\label{T:DSC_contrast}

\end{table}

In~\cite{sharma2017automatic}, 16,000 original annotated images were used for training a VGG-16 FCN for semantic segmentation of kidneys with a DSC performance of $86\%$. In our experiments, the focus was on using a limited number of annotated images and considering sequential spatial dependency between images along the z-axis. 
For the purpose of semantic segmentation, the FCNs require the entire volume of annotated original images (i.e., 126 images) as input for training and inference. However, the cylindrical transform method enables us to train contemporary classification networks for whole-image classification without the need for a FCN to predict dense outputs for semantic segmentation.

The performance results of FCN-AlexNet~\cite{krizhevsky2012imagenet}, FCN-GoogLeNet~\cite{szegedy2015going}, and FCN-VGG-19~\cite{simonyan2014very} are presented in Table~\ref{T:DSC_contrast}. The VGG-19 pre-trained on ImageNet~\cite{deng2009imagenet} requires square-size input images. Since cylindrical transformed images are of size $1280\times256$, we did not use pre-trained models. However, we used FCN-VGG-19~\cite{simonyan2014very} for training with original images for sake of comparison. The experiments were conducted in five schemes: 1)~from scratch end-to-end in an FCN mode; 2)~using pre-trained weights (denoted with P in the tables) on ImageNet~\cite{deng2009imagenet} end-to-end in a FCN mode; 3)~from scratch end-to-end in an FCN mode with augmentation; 4)~using pre-trained weights on ImageNet~\cite{deng2009imagenet} end-to-end in an FCN mode with augmentation; 5)~from scratch using cylindrical transformed (denoted with CT in the tables) images. The augmentation methods used in the FCNs include rotation (every 36 degrees - 10$\times$), scaling (${\{0.7,1.3\}}$ - 2$\times$), shifting an image in x-y direction (${\{\pm50,\pm100\}}$ - 2$\times$), and applying an intensity variation (${\{0.5,1.5\}}$ - 2$\times$) similar to~\cite{sharma2017automatic}, totaling ${126\times80=10,080}$ training images.

\begin{figure}[!t]
\vspace{-4mm}
\centering
\captionsetup{font=small}
 \begin{subfigure}[t]{0.25\textwidth}
        \centering
                \includegraphics[width=1.0\textwidth]{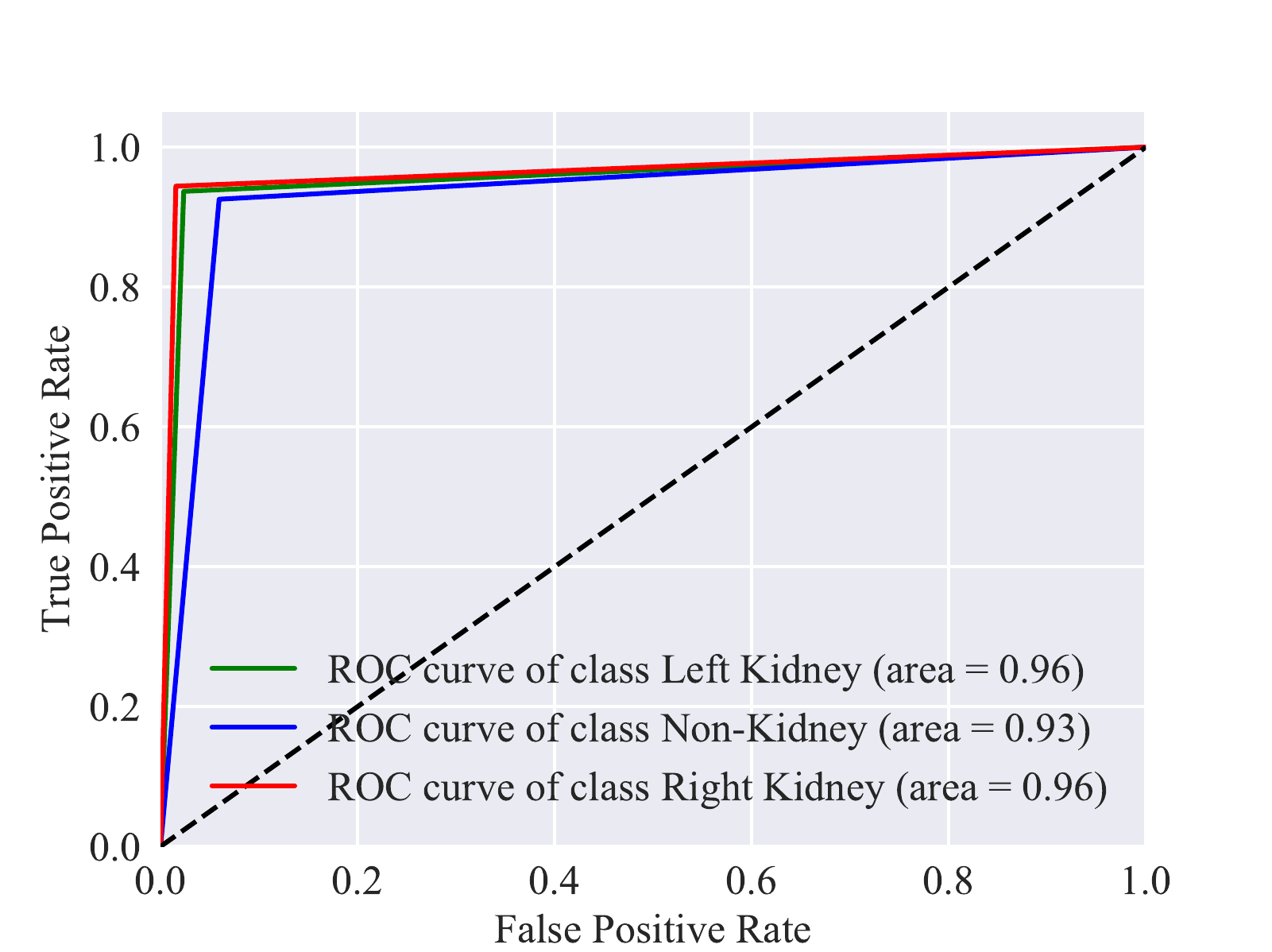}
                \caption{AlexNet.}
                \label{fig:}
        \end{subfigure}%   
        \hspace{-4mm} 
           \begin{subfigure}[t]{0.25\textwidth}
        \centering
                \includegraphics[width=1.0\textwidth]{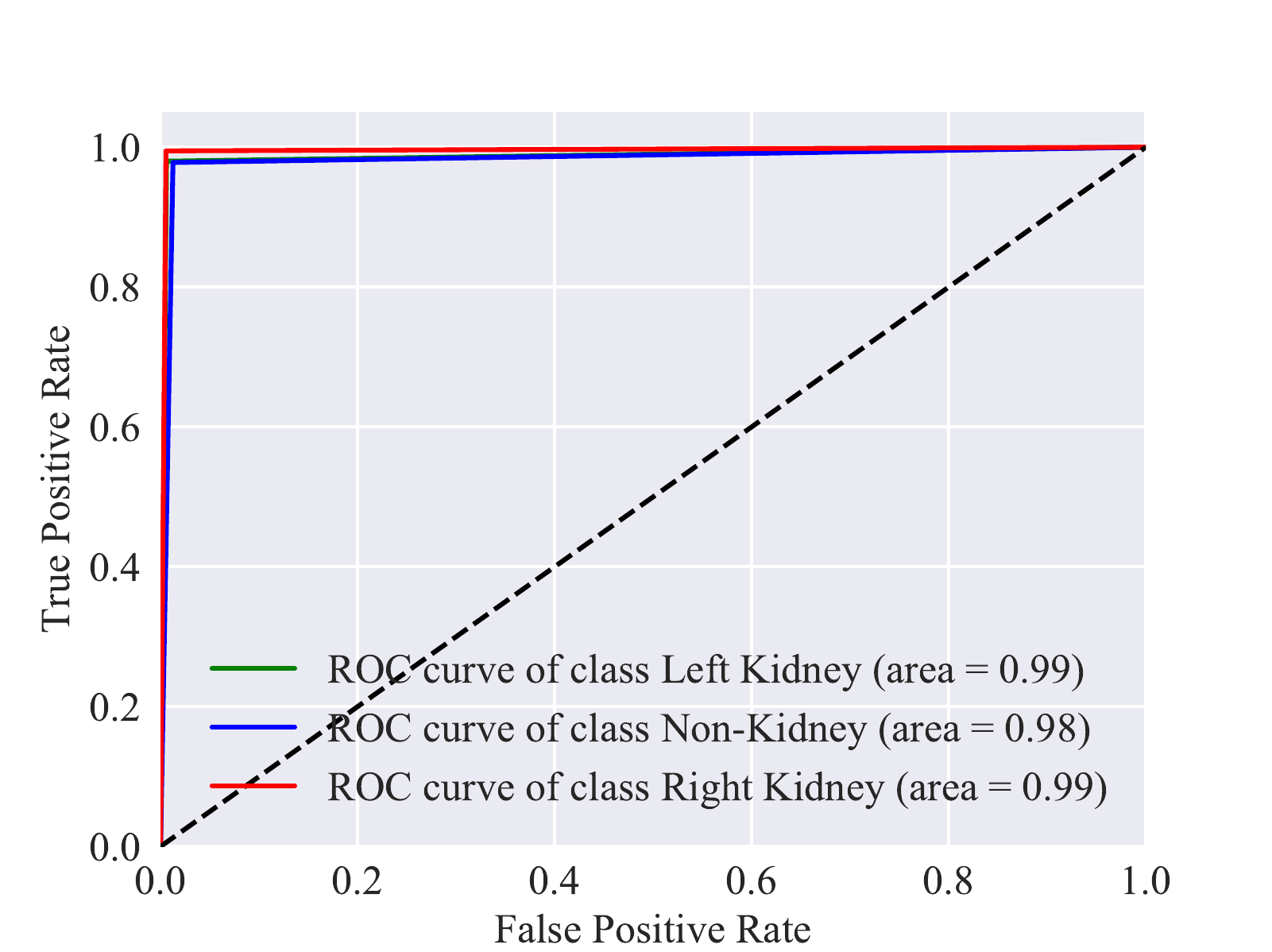}
                \caption{GoogLeNet.}
                \label{fig:}
        \end{subfigure}%                
        \caption{ROC curve and AUC of AlexNet and GoogLeNet on the test dataset for contrast CTs.}       
\label{fig:roc_curves}
\vspace{-2mm}
\end{figure}

Table~\ref{T:f1score_non_contrast} shows precision and recall scores of the DCNNs evaluated in Table~\ref{T:DSC_contrast}. The receiver operating characteristic (ROC) plots in Figure~\ref{fig:roc_curves} show the area under curve (AUC) of the classification models trained using cylindrical transformed images, which is $95.00\%$ and $98.66\%$ for AlexNet and GoogLeNet, respectively. The overall performance results show that FCNs are challenging to train with a limited number of training images. These models have achieved less DSC performance comparing to the GoogLeNet, trained with cylindrical transform generated images, that produced a DSC value of $98.40\%$.

\begin{table}[]
\centering
\captionsetup{font=footnotesize}
\caption{Precision and recall of cylindrical transform (CLT) method for contrast abdominal CT.}
\label{T:f1score_non_contrast}
\begin{adjustbox}{width=0.3\textwidth}
\begin{tabular}{|c|c|c|c|c|c|}
\hline
\multirow{5}{*}{Model} & \multirow{5}{*}{Measure} & \multicolumn{3}{c|}{ Class} & \multirow{5}{*}{Avg.} \\ \cline{3-5}

&  &\rotatebox[origin=c]{90}{Left Kidney}  & \rotatebox[origin=c]{90}{Non-Kidney} & \rotatebox[origin=c]{90}{Right Kidney}  &   \\ \hline \hline

%\multirow{3}{*}{FCN-GoogLeNet} & precision &  &  &  &  \\ \cline{2-6} 
% & recall &  &  &  &  \\ \cline{2-6} 
% & f1-score &  &  &  &  \\ \hline
%\multirow{3}{*}{FCN-AlexNet} & precision &  &  &  &  \\ \cline{2-6} 
% & recall &  &  &  &  \\ \cline{2-6} 
% & f1-score &  &  &  &  \\ \hline
\multirow{3}{*}{\begin{tabular}[c]{@{}c@{}}CLT-AlexNet\end{tabular}} &precision &0.95 & 0.90 & 0.97 & 0.94 \\ \cline{2-6} 
 & recall &  0.94 & 0.93 & 0.94 & 0.94 \\ \cline{2-6}  
 & f1-score & 0.94 & 0.91 & 0.96 & 0.94\\ \hline \hline

\multirow{3}{*}{\begin{tabular}[c]{@{}c@{}}CLT-GoogLeNet\end{tabular}} & precision &0.99 & 0.98 & 0.99 & $\mathbf{0.98}$\\ \cline{2-6} 
 & recall &  0.98 & 0.98 & 0.99 & $\mathbf{0.98}$\\ \cline{2-6} 
 & f1-score & 0.98 & 0.98 & 0.99 & $\mathbf{0.98}$\\ \hline

%\multirow{2}{*}{\begin{tabular}[c]{@{}c@{}}RT-VGG19\end{tabular}} & precision &0.35	&0.79	&0.45		&0.53\\ \cline{2-6} 
% & recall &0.17&	0.29	&1.00	&	0.49\\ \hline\hline
% 
%% \cline{2-10} 
%% & f1-score & 0.16	&0.36	&0.62&	0.46 &0.23&	0.42	&0.62		&0.51\\ \hline\hline
%
%\multirow{2}{*}{\begin{tabular}[c]{@{}c@{}}RT-VGG19P\end{tabular}} & 
% precision &0.99	&0.99&	0.99	&	$\mathbf{0.99}$\\ \cline{2-6} 
% & recall &0.99	&0.98	&0.99	&	$\mathbf{0.99}$\\ \hline
 
% \cline{2-10} 
% & f1-score &0.95	&0.95	&0.98&	$\mathbf{0.96}$&0.99&	0.98	&0.99	&	$\mathbf{0.99}$\\ \hline\hline
\end{tabular}
\end{adjustbox}
%\vspace{-mm}
\end{table}

\section{Conclusions}
\label{sec:conclusion}
%Training supervised deep neural networks for semantic segmentation requires large volume of annotated images. Annotation of images is a expensive task, particularly in medical imaging where the quality of annotation is very important and due to privacy issues sharing of the images is challenging. In this paper, we propose a novel semantic segmentation method for training deep neural networks, which requires very little annotated images.
Most of the proposed methods for semantic segmentation of sequential images (i.e., a volume) perform segmentation for each image of the sequence independently, without considering the sequential spatial dependency between the images. In addition, annotating sequential images is challenging and expensive, which is a drawback in using supervised deep learning models due to their need for a massive number of training samples. In this paper, we investigate the semantic segmentation of sequential images in a 3D space by proposing a sampling method in the cylindrical coordinate system. The proposed method can generate images up to the number of pixels in the volume, and therefore augment the training dataset. The generated images contain spatial samples from the x-y plane, as well as the time (i.e., sequential) dimension along the z-axis. This method enables us to train contemporary classification convolutional neural networks instead of a fully convolutional network (FCN). This technique helps the network to avoid overfitting and boost up its generalization performance. 

%For future work, this method can be used for semantic segmentation of other sequential modalities in medical imaging as well as other time-series applications.

\bibliographystyle{IEEEtran}
\bibliography{CTLIEEEtrans,mybibfile}

\end{document}